%% file: main.tex
\documentclass{article}

\usepackage{PRIMEarxiv}

\usepackage[utf8]{inputenc} 
\usepackage[T1]{fontenc}    
\usepackage{hyperref}       
\usepackage{url}            
\usepackage{booktabs}       
\usepackage{amsfonts}       
\usepackage{nicefrac}       
\usepackage{microtype}      
\usepackage{lipsum}
\usepackage{fancyhdr}       
\usepackage{graphicx}       
\usepackage{subcaption}
\usepackage{xspace}
\usepackage{titlesec}
\usepackage{makecell}
\usepackage{multirow}

\graphicspath{{media/}}     

\title{Correlation-Driven Multi-Level Multimodal Learning for Anomaly Detection on Multiple Energy Sources
\thanks{\textit{\underline{Citation}}: 
\textbf{Authors. Title. Pages.... DOI:000000/11111.}} 
}

\author{
  Taehee Kim \\
  Graduate School of Data Science \\
  Seoul National University of Science and Technology \\
  Seoul, South Korea\\
  \texttt{cltkim@seoultech.ac.kr} \\
   \And
  Hyuk-Yoon Kwon \\
  Graduate School of Data Science \\
  Seoul National University of Science and Technology \\
  Seoul, South Korea\\
  \texttt{hyukyoon.kwon@seoultech.ac.kr} \\
}

\begin{document}
\maketitle

\begin{abstract} Advanced metering infrastructure~(AMI) has been widely used as an intelligent energy consumption measurement system. Electric power was the representative energy source that can be collected by AMI; most existing studies to detect abnormal energy consumption have focused on a single energy source, i.e., power. Recently, other energy sources such as water, gas, and heating have also been actively collected. As a result, it is necessary to develop a unified methodology for anomaly detection across multiple energy sources; however, research efforts have rarely been made to tackle this issue. The inherent difficulty with this issue stems from the fact that anomalies are not usually annotated. Moreover, existing works of anomaly definition depend on only individual energy sources. In this paper, we first propose a method for defining anomalies considering not only individual energy sources but also correlations between them. Then, we propose a new Correlation-driven Multi-Level Multimodal Learning model for anomaly detection on multiple energy sources. The distinguishing property of the model incorporates multiple energy sources in multi-levels based on the strengths of the correlations between them. Furthermore, we generalize the proposed model in order to integrate arbitrary new energy sources with further performance improvement, considering not only correlated but also non-correlated sources. Through extensive experiments on real-world datasets consisting of three to five energy sources, we demonstrate that the proposed model clearly outperforms the existing multimodal learning and recent time-series anomaly detection models, and we observe that our model makes further the performance improvement as more correlated or non-correlated energy sources are integrated.

\end{abstract}


\input{ch1_introduction.tex}

\input{ch2_related_work}

\input{ch3_dataset.tex}
\input{ch4_background.tex}

\input{ch5_proposed_model.tex}
\input{ch6_performance_evaluation.tex}
\input{ch7_conclusion.tex}

\section*{Acknowledgments}
This work was supported by the National Research Foundation of Korea(NRF) grant funded by the Korea government(MSIT) (No. 2022R1F1A1067008), and by the Basic Science Research Program through the National Research Foundation of Korea(NRF) funded by the Ministry of Education (No. 2019R1A6A1A03032119)

\bibliographystyle{unsrt}  
\bibliography{references}

\end{document}

%% file: ch1_introduction.tex
\section{INTRODUCTION}

Advanced Metering Infrastructure~(AMI) is an intelligent system for measuring energy consumption and is used to provide useful information such as time-varying usage rates, electricity savings, and demand reactions~\cite{yang2020you, maamar2018machine, himeur2021artificial, chou2014real, korba2020anomaly, anupong2022towards}. Its importance is emerging in line with the government's policy of establishing smart grids and reducing costs by improving energy efficiency~\cite{himeur2021artificial}. Various studies have shown the importance of detecting abnormal energy consumption patterns. Chou et al.~\cite{chou2014real} stated that energy consumption required for the operation and maintenance of buildings could be reduced by discovering abnormal patterns in energy consumption. Korba et al.~\cite{korba2020anomaly} described that detecting abnormal energy consumption is core for further savings.

Electric power was the most representative energy source collected through AMI; most existing studies to detect anomalous energy consumption have focused on one specific energy source, i.e., power. Recently, other energy sources such as water, gas, and heating have also been actively collected~\cite{anupong2022towards}. Accordingly, multi-energy systems~(MES) have also brought new attention to energy consumption analysis on multiple-energy sources, i.e., power, heating, cooling, fuels, and gas, for both operational and planning stages~\cite{mancarella2014mes}. However, they have been focused on the costs and benefits of policy initiatives, but have rarely been studied for anomaly detection of multiple energy sources. Therefore, we note that developing a unified anomaly detection methodology on multiple energy sources is worthwhile.

However, anomaly definition is challenging because anomalies are not usually labeled, and the criteria for defining them are not clear by energy consumption alone~\cite{weng2018multi}. To address lack of the annotated data, previous studies have defined anomalies by using regression-based approaches~\cite{weng2018multi, zhang2011anomaly, wang2018identifying} or existing anomaly detection algorithms~\cite{weng2018multi, malhotra2016lstm, li2019specae}. In this paper, we present a new approach for defining anomalous energy consumption not only by individuals but also based on the correlations between energy sources. In particular, We note that abnormal observations are detected by the correlations between energy sources, although these might be considered to be normal by a single energy source.

\begin{figure}[!t]
\centerline{\includegraphics[width=0.56\textwidth]{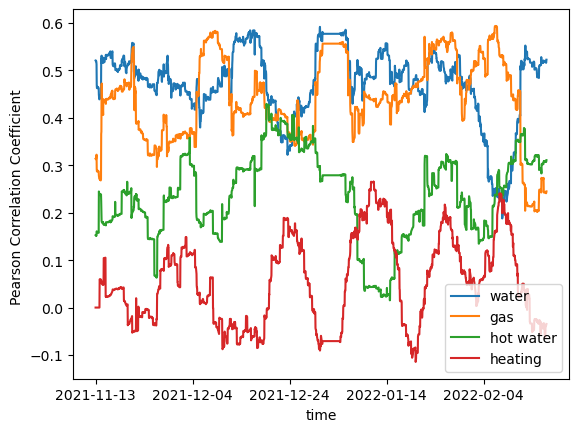}}
\caption{Pearson correlation coefficients between power and other energy consumption.}
\label{correlations}
\end{figure}

Fig.~\ref{correlations} shows the correlation of the consumption between power and other energy sources, i.e., water, gas, hot water, and heating on a weekly basis using a real-world dataset, 5-ECK-2022, which is used in the experiments. This indicates that significant correlations are observed between multiple energy sources consumed in a household. We also note that the degree of the correlations among energy sources is variable. Specifically, the average Pearson correlation coefficient between power and water consumption is 0.4726, indicating that they are strongly correlated; however, that between power and heating consumption is 0.0539, indicating that they are not correlated. Based on these observations, correlations among energy sources need to be considered in order to detect abnormal consumption, instead of considering only one single energy source. Moreover, we need to reflect them discriminately according to the strength of the correlations between energy sources. 

\begin{figure}[!t]
\centerline{\includegraphics[width=0.6\textwidth]{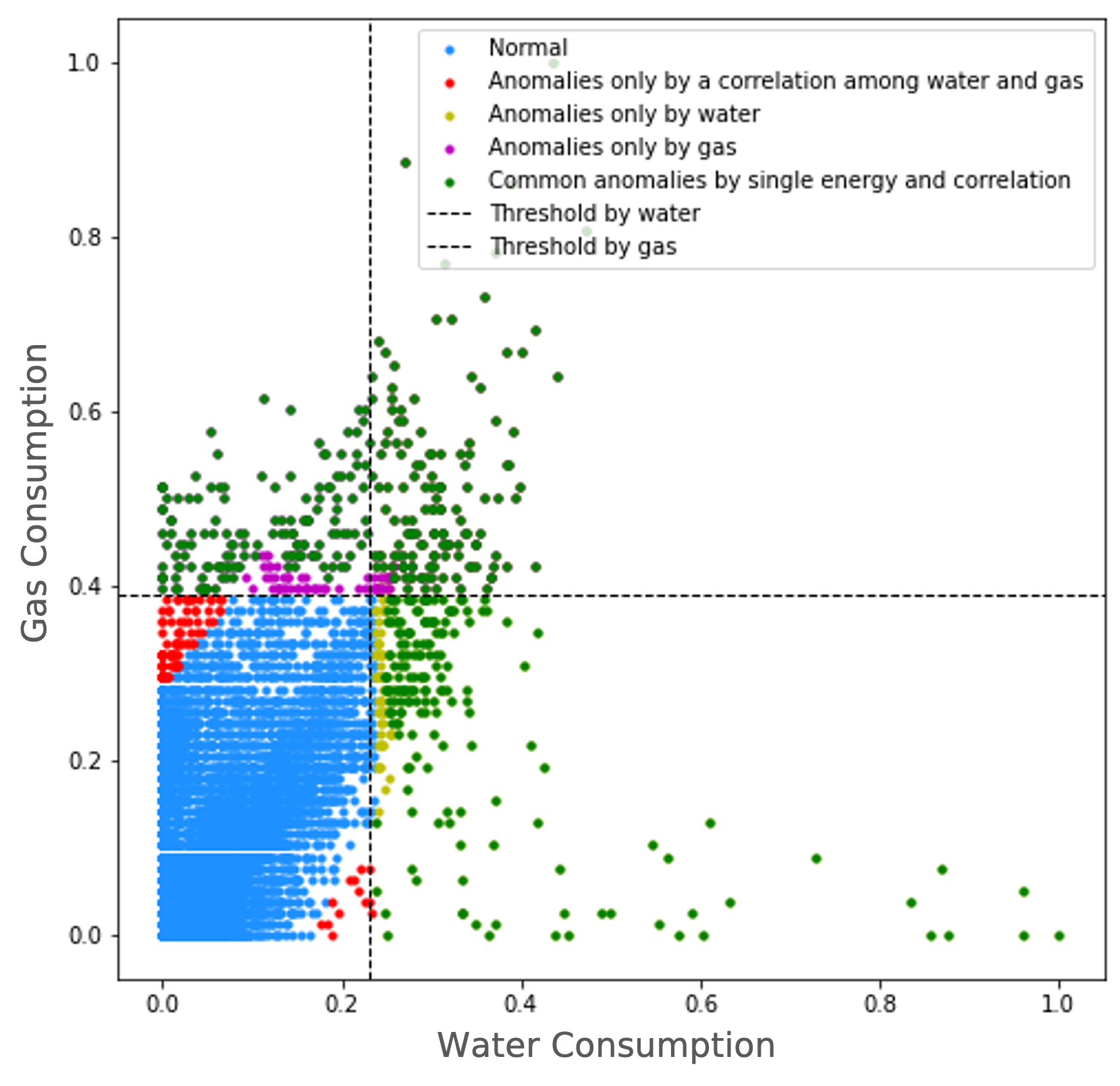}}
\caption{A distribution of anomalies defined by a single source and the correlation between energy sources.}
\label{distribution}
\end{figure}

Fig.~\ref{distribution} illustrates anomalies defined by two different approaches on a real-world energy consumption dataset of gas and water: 1)~anomalies defined by the correlation between them and 2)~anomalies defined by a single energy source based on the method that has been primarily used in previous studies~\cite{weng2018multi, wang2018identifying}. In this figure, vertical and horizontal dotted lines show the criteria to separate normal and abnormal consumption by each single energy source. We need to pay attention to anomalies in red dots defined by the correlation between gas and water consumption, which are not anomalies by each energy source. It is worthwhile to consider them as anomalies because unnatural relationships of the consumption for correlated energy sources also imply abnormal patterns. Therefore, we present an anomaly definition method that augments the anomalies by the correlations between energy sources, which ensures the mutual relationship between energy sources, to the anomalies defined by each single energy source. In this paper, we present a method for anomaly definition to define anomalies based on correlations between multiple energy sources using Cook's distance~\cite{cook1977detection}, identifying influential instances based on the relationship.

We propose a new multimodal learning model to detect anomalies by the strength of the correlations among multiple energy sources. As the simplest model architecture, we consider three kinds of energy sources, which have been commonly used for multiple energy sources, and design them in two-level employing different fusion methods~(i.e., early fusion and late fusion) based on the correlation between energy sources. Specifically, we integrate two energy sources that have weaker correlations than the other by early fusion~(i.e., fusing the raw data) in the \textit{first-level fusion layer}, and then integrate the fused result with a remaining energy source by late fusion~(i.e., fusing the features extracted from raw data) in the \textit{second-level fusion layer}.

We further provide a scalable model architecture capable of effectively extending to more energy sources by maintaining a correlation-driven approach for the energy sources that are correlated with each other. Basically, we separate the correlated energy sources into weakly and strongly correlated sources and integrate the former with early fusion in the first-level layer and the latter with late fusion in the second-level layer. Specifically, we first choose two energy sources that have the minimum correlation with the other energy sources for early fusion. Then, out of the remaining energy sources, we choose the next energy source that has the weakest correlation with the prior two energy sources. This source is integrated for early fusion only when its average of the Pearson correlation coefficient with two energy sources composed of early fusion is less than a correlated threshold; this is repeated for all the remaining energy sources. Otherwise, the remaining sources that are not chosen for early fusion are integrated with late fusion.

Furthermore, for the generality of the proposed model, we devise a model architecture to integrate non-correlated energy sources, which possibly occurs in practice when dealing with multiple energy sources, with the correlated energy sources in the \textit{third-level fusion layer}. The basic strategy is that we completely separate non-correlated energy sources from the integrated multimodal learning model for correlated energy sources, where we perform feature extraction for each non-correlated energy source individually. Then, in the third-level fusion, we concatenate the results of early fusion, those of late fusion, and the features extracted from the non-correlated energy sources in a supervised manner. Here, by training the third-level fusion layer with the datasets where anomalies are defined by considering the correlation between energy sources, we completely differentiate weight contributions between correlated and non-correlated energy sources, which have been confirmed by the SHAP values.

Through extensive experiments on three real-world datasets that contain various types of multiple energy sources, i.e., from three to five sources, we demonstrate the effectiveness of the proposed model. First, we compare the proposed model with the existing multimodal learning models and recent time-series anomaly detection models. For the datasets consisting of three energy sources, i.e., AMPds2, 3-ECK-2021, and 3-ECK-2022 where all the sources are correlated with each other, our model outperforms the existing models, improving the F1 score by 0.56\%$\sim$26.85\%. Second, for the dataset consisting of four correlated energy sources, 4-ECK-2022, our model maintains the performance improvement over the existing models. We also confirm the scalability of our method by showing that the performance improvement of the average F1 score compared to the existing models is 13.73$\sim$14.88\% on 4-ECK-2022, whereas it was 2.02$\sim$13.16\% on 3-ECK-2022. Finally, on the datasets where a non-correlated energy source is integrated, our model improves the average F1 score of the existing models by 3.25$\sim$23.36\%, whereas it was 2.67$\sim$14.88\% on the same dataset only excluded a non-correlated energy source. 

The main contributions of this paper are summarized as follows.
\begin{enumerate}
    \item We devise a new anomaly definition method on energy consumption consisting of multiple energy sources. In particular, we consider not only individual energy sources but also their correlation with energy consumption. Therefore, we can define abnormal energy consumption by considering the mutual relationships between correlated energy sources even if each of them can be normal with respect to a single energy source.
    
    \item We propose a novel multimodal learning model to detect anomalies in energy consumption, considering multiple energy sources. The salient point of the proposed model adjusts the order of fusing energy sources based on the correlation among them in the model, resulting in a multi-level architecture. 

    \item The proposed model is scalable in integrating more arbitrary energy sources. We formalize criteria for integrating each energy source one by one into a unified model based on the correlation between energy sources. We confirm that the performance improvement of our model becomes greater as the energy sources increase. 

    \item The proposed model can effectively deal with non-correlated energy sources by separately dealing with each of them from a unified multimodal learning model for correlated energy sources. Then, we differentiate the weight contribution between correlated and non-correlated sources by training them with annotated datasets defined by considering the correlation between energy sources. Due to this novel feature, the performance of our model is further improved when non-correlated energy sources are added to the correlated energy sources. 
\end{enumerate}

The rest of this paper is organized as follows. In Section~\ref{related-work}, we describe the related work. In Section~\ref{dataset}, we explain real-world energy consumption datasets consisting of multiple energy sources. In Section~\ref{background}, we describe the background of the proposed model. In Section~\ref{proposed-method}, we propose a new multimodal learning model for anomaly detection. In Section~\ref{experiments}, we describe the performance evaluation. In Section~\ref{conclusion} we conclude the paper.

%% file: ch2_related_work.tex
\section{RELATED WORK}\label{related-work}

\subsection{Anomaly on Energy Consumption}

To resolve the lack of the annotated anomalous energy consumption in the dataset, some previous studies have defined anomalies from energy consumption datasets. Zhang et al.~\cite{zhang2011anomaly} presented a linear regression-based method between power consumption and temperature. Wang et al.~\cite{wang2018identifying} identified anomalies using the Mahalanobis distance~\cite{taguchi2000mahalanobis} between gas consumption and temperature. Cook’s distance~\cite{cook1977detection} and Different-in-Fits (DFFITS)~\cite{belsley2005regression} were also used to measure leverage-based anomaly scores. Weng et al.~\cite{weng2018multi} defined anomalies by combining the results of the existing anomaly detection algorithms (i.e., isolation forest and feature bagging).

In this paper, we present a method to define anomalies by correlations between energy sources instead of using only individual sources. The correlation-based anomaly definition takes advantage of considering the relationships between energy sources, which means we can utilize consumption patterns between correlated energy sources for detecting abnormal energy consumption. For example, in Fig. 1, red dots are defined as normal by individual energy consumption, but as abnormal by correlation-based consumption. These types of anomalies can detect important abnormal consumption patterns such as leakage or theft of a specific energy source. \

\subsection{Anomaly Detection for a Single Energy Source}

Most of the existing methods for detecting anomalies in energy consumption have been conducted for electric power~\cite{maamar2018machine, himeur2021artificial, chou2014real, korba2020anomaly, anupong2022towards, madhure2020cnn, peng2021electricity}. Madhure et al.~\cite{madhure2020cnn} proposed a CNN-LSTM-based model and Peng et al.~\cite{peng2021electricity} used \textit{K}-means clustering and local outlier factor to detect electricity theft. There have been a few studies on other energy sources, i.e., water~\cite{yu2021study, iyer2019blockchain}, gas~\cite{yang2020you, de2015short}, heating~\cite{lumbreras2021unsupervised, park2020explainable}. Iyer et al.~\cite{iyer2019blockchain} detected potential fraud with respect to wastewater reuse; Nadai et al.~\cite{de2015short} identified the causes of gas wastage; Park et al.~\cite{park2020explainable} detected a heating system malfunction. 

\subsection{Anomaly Detection for Multiple Energy Sources}

Only limited studies have been conducted for detecting anomalies on multiple energy sources. Weng et al.~\cite{weng2018multi} proposed an unsupervised anomaly detection scheme for energy consumption for power, water, and gas sources. They applied LSTM AutoEncoder (LSTM-AE)~\cite{malhotra2016lstm} to each of multiple energy sources, comparing the proposed model with the existing linear models (i.e., PCA and one-class support vector machine), proximity-based model (i.e., k-nearest neighbors), and ensemble models using isolation forest and feature bagging. Although they considered several energy sources, each source was treated independently and unified anomaly detection considering multiple energy sources was not considered. Given that energy consumption is highly correlated with multiple energy sources, anomaly detection considering multiple energy sources in integrated consumption is crucial because it enables us to detect anomalies that cannot be detected only by a specific energy source.

\subsection{Anomaly Detection for Time-Series Data}

  Energy consumption can be regarded as a kind of time-series data. Research efforts for anomaly detection on time-series data have been widely performed. Anomaly detection in time-series data is classified into three categories: 1)~clustering-based, 2)~machine learning-based, and 3)~neural network-based approaches. Clustering-based approaches employed \textit{k}-means~\cite{kiss2014data, ccelik2011anomaly} and DBSCAN~\cite{yu2021study, ccelik2011anomaly}; Machine learning-based approaches utilized isolation forest~\cite{weng2018multi}, \textit{k}-nearest neighbors~\cite{huang2019identification}, one-class support vector machines~\cite{tao2020kernel, budiarto2019unsupervised}, and local outlier factor~\cite{peng2021electricity, budiarto2019unsupervised}. Recently, neural network-based methods have outperformed the existing clustering or machine learning-based approaches~\cite{weng2018multi, bashar2020tanogan}. Malhotra et al.~\cite{malhotra2016lstm} proposed LSTM-AE to minimize reconstruction error that occurs during a training process. Li et al.~\cite{li2019specae} proposed a graph convolution and deconvolution-based AutoEncoder for anomaly detection in attributed networks. Goodfellow et al.~\cite{goodfellow2020generative} improved anomaly detection performance on time-series data by adversarial training. Li et al.~\cite{li2019mad} proposed MAD-GAN that utilizes a recurrent neural network, i.e., LSTM-RNN, as GAN’s generator and discriminator. Audibert et al.~\cite{audibert2020usad} proposed USAD that trains adversarial models within an AutoEncoder architecture, showing that it outperforms the existing models including AutoEncoder, isolation forest, deep autoencoding gaussian mixture model, OmniAnomaly, and LSTM-VAE. 

Even if multiple energy sources are a kind of multivariate time series, they have distinguishing properties: they measure the same household’s energy consumption, and they have distinguishing correlation patterns based on the relationship between energy sources. Therefore, in this paper, we aim to devise a customized model that emphasizes the difference in the correlations among multiple energy sources.

\subsection{Multimodal Learning}

Multimodal learning incorporates a variety of modalities that have distinct characteristics, such as images, texts, and sensors, in a learning model~\cite{hu2021detection, baltruvsaitis2018multimodal}. It has been widely applied to various domains to enhance the performance of unimodal learning. Nemati et al.~\cite{nemati2017exploiting} demonstrated the effects of combining two types of textual features (i.e., subtitles and song lyrics) with the audio-visual contents of music videos for video retrieval. Poria et al.~\cite{poria2016fusing} proposed a method for sentiment analysis using features derived from audio, visual, and textual modalities. 
Neverova et al.~\cite{neverova2015moddrop} utilized multiple modalities such as depth, grayscale video, pose signals, and audio for gesture detection. Park et al.~\cite{park2022multimodal} proposed a multimodal learning model integrating various features extracted from video, audio, and text to improve the performance of detecting highlights of e-sports videos. 
Li et al.~\cite{8966989} proposed a multimodal learning model incorporating news information and stock fundamentals to predict stock price by resolving their heterogeneity. Shin et al.~\cite{shin2020new} proposed a contrastive word embedding model to improve the performance of deep learning-based classification models for detecting cyber-security relevant tweets by separating the entire word corpus into positive and negative ones and fusing the extracted features from them by a multimodal learning model. Wang et al.~\cite{9470948} utilized multi geographic granules from city-level to mile-level in a multimodal learning to improve the forecasting of housing price.

%% file: ch3_dataset.tex
\section{DATASET}\label{dataset}
To cover various energy consumption datasets with multiple energy sources, we used three real-world datasets: 1) AMPds2, 2) 4-ECK-2021~(Four types of Energy Consumption in Korea), and 3) 5-ECK-2022~(Five types of Energy Consumption in Korea). Based on them, we reconstruct more datasets to include only correlated energy sources for extensive comparisons of the performance between datasets. We define an energy source \textit{ES} as correlated when its Pearson correlation coefficient with at least one another energy source is larger than $Th_{correlated}$ and as non-correlated when its Pearson correlation coefficient with every other energy source is less than $Th_{correlated}$. In this paper, we use 0.2 for $Th_{correlated}$ followed by the previous study~\cite{akoglu2018user}. For preprocessing of datasets, we constantly transformed raw datasets to make the measurement frequency as an hour, and missing values were filled based on linear interpolation between the consecutive segments. Then, min-max scaling was used for normalization. \\

\noindent \textbf{The Almanac of Minutely Power dataset (AMPds2)}\footnote{https://doi.org/10.7910/DVN/FIE0S4}: This dataset includes three types of energy sources: 1) power, 2) gas, and 3) water, collected from a single household in Canada. This dataset includes 1,051,200 records observed at minute intervals from Apr. 2012 to Mar. 2014. \\

\noindent \textbf{Four types of Energy Consumption in Korea (4-ECK-2021)}\footnote{https://doi.org/10.17632/m68xz4w4t9.2}: This dataset includes four types of energy sources: 1) power, 2) water, 3) hot water, and heating. It has 47,117 records observed at 15-minute intervals from Jan. 2020 to Mar. 2021. To contain only correlated energy sources, we redefine 3-ECK-2021 by excluding a non-correlated energy source, heating, from 4-ECK-2021.\\

\noindent \textbf{Five types of Energy Consumption in Korea (5-ECK-2022)}\footnote{https://doi.org/10.17632/m68xz4w4t9.2}: This dataset includes five types of energy sources: 1) electricity, 2) gas, 3) water, 4) hot water, and 5) heating. It has 13,584 records observed at one hour intervals from Aug. 2020 to Feb. 2022. We redefine 3-ECK-2022 and 4-ECK-2022 by excluding non-correlated energy sources from 5-ECK-2022. Specifically, a heating energy source, which has the weakest correlation with the other energy sources, is excluded from 5-ECK-2022 and defined as 4-ECK-2022; heating and hot water are excluded from 5-ECK-2022 and defined as 3-ECK-2022.

\begin{figure}[!t]
\captionsetup[subfigure]{justification=centering}
     \centering
     \begin{subfigure}[h]{0.33\textwidth}
         \centering
         \includegraphics[width=\textwidth]{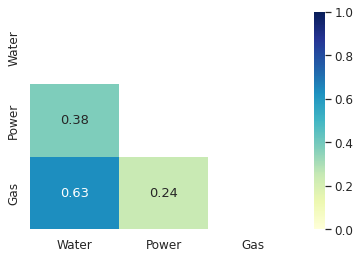}
         \caption{AMPds2}
         \label{dataset-a}
     \end{subfigure} 
     \hfill
     \begin{subfigure}[h]{0.33\textwidth}
         \centering
         \includegraphics[width=\textwidth]{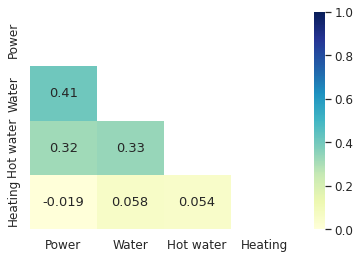}
         \caption{4-ECK-2021}
         \label{dataset-b}
     \end{subfigure}
     \hfill
     \begin{subfigure}[h]{0.33\textwidth}
         \centering
         \includegraphics[width=\textwidth]{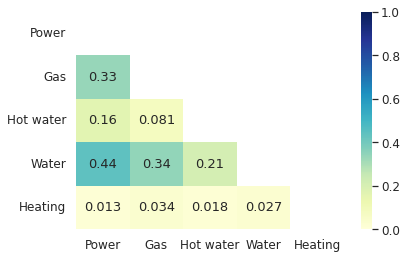}
         \caption{5-ECK-2022}
         \label{dataset-c}
     \end{subfigure}
        \caption{Pearson correlation coefficients between multiple energy sources.}
        \label{dataset-correlations}
        \vspace*{-0.2cm}
\end{figure}

Fig.~\ref{dataset-correlations} shows a correlation matrix between energy sources based on the Pearson correlation coefficient for each dataset: Fig.~\ref{dataset-a} for AMPds2; Fig.~\ref{dataset-b} for 4-ECK-2021; Fig.~\ref{dataset-c} for 5-ECK-2022. Overall, the common characteristics are observed across the datasets, and distinct characteristics are also observed depending on the datasets. Specifically, in AMPds2, water is the most strongly correlated with the other sources; In 4-ECK-2021 and 5-ECK-2022, both power and water are strongly correlated with the others, while heating is overall relatively less correlated with the others.

%% file: ch4_background.tex
\section{BACKGROUND}\label{background}
\subsection{Multimodal Learning}

\begin{figure}[h]
\captionsetup[subfigure]{justification=centering}
     \centering
     \begin{subfigure}[h]{0.38\textwidth}
         \includegraphics[width=\textwidth]{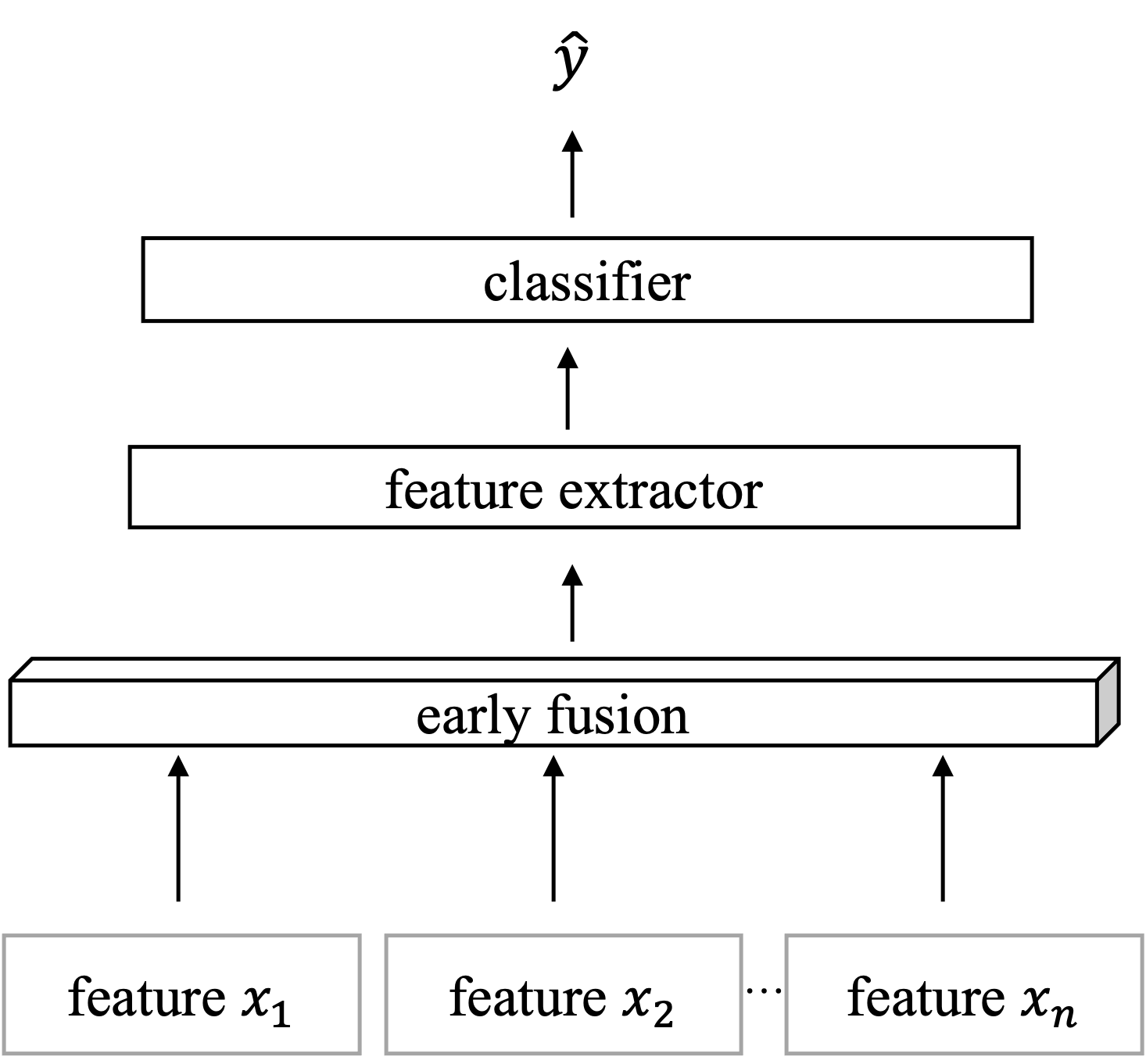}
         \caption{Early fusion}
         \label{earlyfusion}
     \end{subfigure} 
    \hfill
     \begin{subfigure}[h]{0.42\textwidth}
         \centering
         \includegraphics[width=\textwidth]{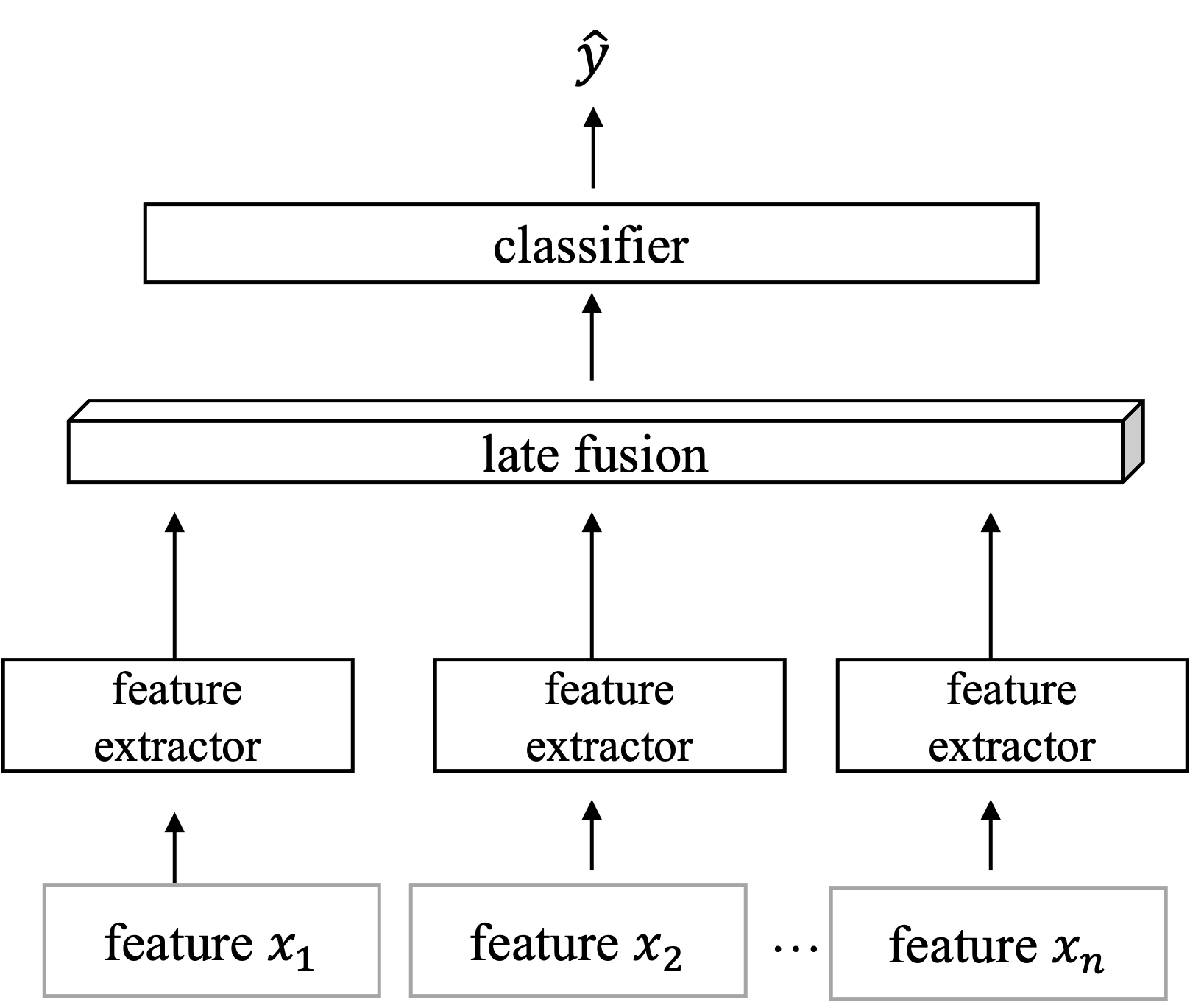}
         \caption{Late fusion}
         \label{latefusion}
     \end{subfigure}
    \caption{Fusion methods for multimodal learning.}
    \label{fusionmethods}
\end{figure}

Fig.~\ref{fusionmethods} shows two categories for the multimodal learning models: 1)~early fusion and 2)~late fusion\cite{baltruvsaitis2018multimodal, atrey2010multimodal}. Early fusion concatenates raw features at a low level. A feature extractor is applied to the integrated features, and then the final classifier is applied. In contrast, in late fusion, a separate feature extractor is applied to each feature. Their results are integrated, and these embedded results are fed into the final classifier. According to the fusion strategies, we note that they integrate multiple features in one level. In this study, we devise a multimodal learning model in multi-level to reflect the strengths of the correlations differentially.

\subsection{Multivariate Time-series Anomaly Detection}
In this study, we design a model architecture for multimodal learning by using a base anomaly detection model as a basic feature extractor in fusing features. We note that we can employ any existing anomaly detection models for the feature extractor. In this section, we introduce two representative unsupervised models for multivariate time-series anomaly detection, LSTM-AutoEncoder (LSTM-AE)~\cite{malhotra2016lstm} and UnSupervised Anomaly Detection (USAD)~\cite{audibert2020usad}, which are adopted for the feature extractor in this paper. LSTM-AE has been popularly used as an unsupervised learning method for detecting anomalies in multivariate time series. LSTM is used to effectively process sequential data, reflecting long-term temporal dependencies; AutoEncoder plays a role in detecting anomalies based on reconstruction errors that occurred while training. USAD follows an autoencoder architecture where adversarial training is performed in two phases. USAD consists of an encoder and two decoders to take advantage of the autoencoder and adversarial training. In the first phase, the model is trained to reproduce the input data; In the second phase, it is trained to discriminate between real and synthetic reconstructed data created by the first phase. Due to this two-phrase training, the model can effectively detect anomalies even similar to the normal data.

%% file: ch5_proposed_model.tex
\section{PROPOSED MODEL}\label{proposed-method}
\subsection{Definition for Anomalous Energy Consumption}

In this study, we first present a method for anomaly definition in energy consumption to solve the issue, which is the lack of annotated datasets in energy consumption, especially for the dataset composed of multiple energy sources. To this end, following the existing method for anomaly definition based on individual energy sources~\cite{wang2018identifying, weng2018multi}, we augment anomalies by considering the correlation between energy sources. Specifically, we first define anomalies from each energy source based on Z-score~\cite{glantz2001primer}, and we classify three times more than the average consumption as anomalies. Then, we define anomalies by the correlations between energy sources even though they are considered normal in individual energy sources. To measure the abnormality from the energy source to the other based on their correlation, we use Cook's distance~\cite{cook1977detection}, which takes into account both leverage and residuals from each observation. For a non-correlated energy source~(e.g., heating in 4-ECK-2021 and 5-ECK-2022), we define anomalies only based on individual energy sources instead of the correlation-based anomalies.

Cook's distance is defined as in Eq.~\ref{eq1}, which measures the difference between the estimated value by a linear regression model when an instance $i$ is included in the model, $\hat{y_j}$, and the estimated value when $i$ is not included, $\hat{y_{j(i)}}$. Here, $p$ is the number of coefficients, and $MSE$ is the mean squared error.

\begin{equation}\label{eq1}
D_i = \frac{\sum_{j=1}^{n} \hat{y}_{j}- \hat{y}_{j(i)}}{pMSE}
\end{equation}

Cook's distance enables us to examine the relationship between features by measuring the extent to which each instance influences a linear regression model. A high value of $D_i$ means the model is more likely to be distorted by $i$. We use a criterion for Cook's distance to classify the normal and abnormal data as $\frac{4}{N-K-1}$, where $N$ is the number of observations and $K$ is the number of predictors, which has been used in the previous study~\cite{heiberger2015statistical}.

Specifically, let us suppose that we have two correlated energy sources, $CS_p$ and $CS_q$. We can consider two directions in this relationship: 1) from $CS_p$ to $CS_q$ and 2) from $CS_q$ to $CS_p$. As a result, we obtain $D_i$ for each direction by Eq.~(1). Here, we denote the anomalies defined by the former as $A_{sp, sq}$, and anomalies defined by the latter as $A_{sq, sp}$. Then, correlation-based anomalies between $CS_p$ and $CS_q$ are defined as $A_{sp, sq}$ $\cup$ $A_{sq, sp}$. Finally, we merge all correlated-based anomalies for each pair of the correlated energy sources.

\begin{figure}[!t]
\captionsetup[subfigure]{justification=centering}
     \centering
     \begin{subfigure}[h]{0.48\textwidth}
         \centering
         \includegraphics[width=\textwidth]{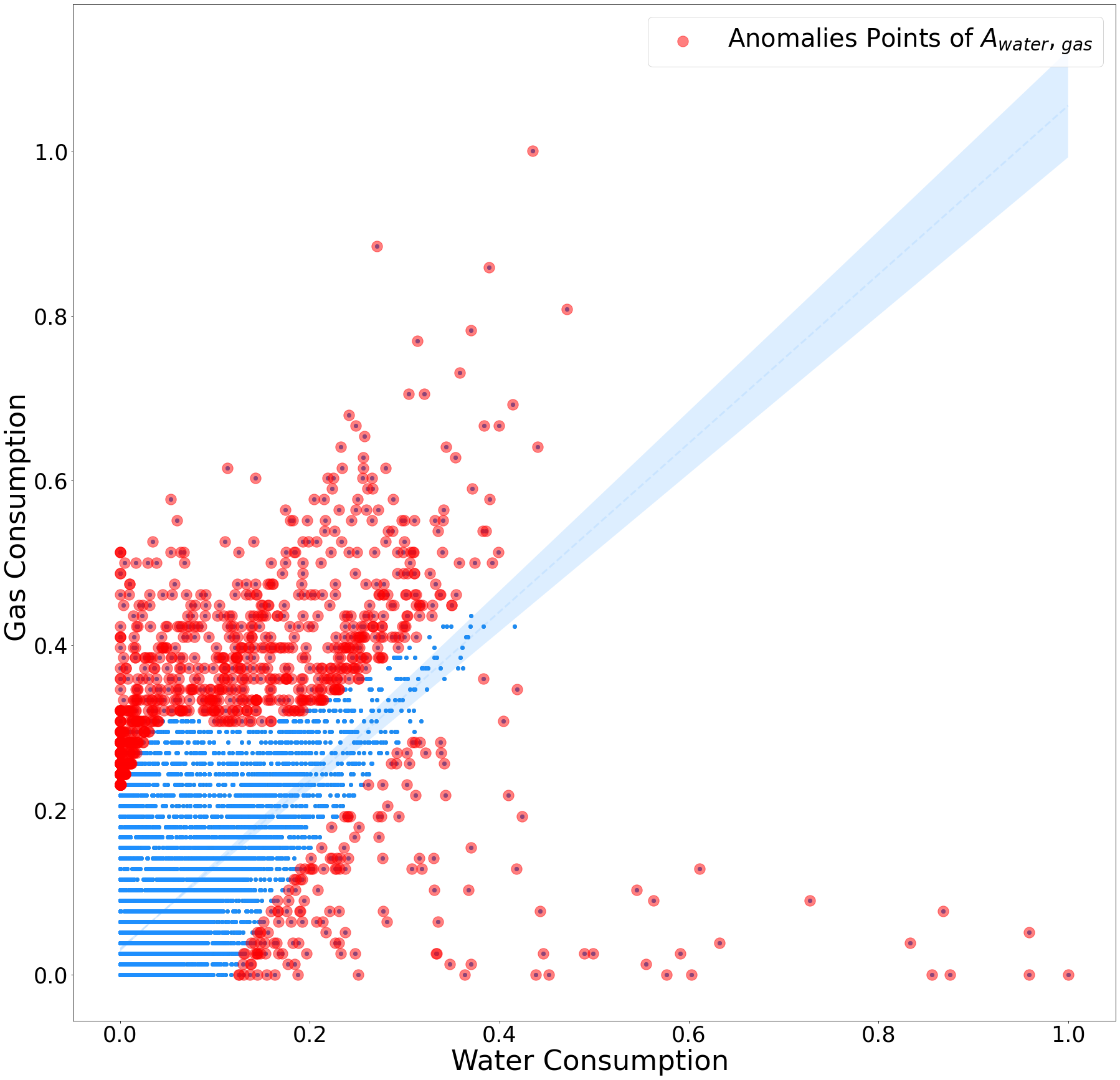}
         \caption{Linear regression of $CS_{gas}$ on $CS_{water}$}
         \label{fig5-a}
     \end{subfigure} 
     \hfill
     \begin{subfigure}[h]{0.48\textwidth}
        \centering
         \includegraphics[width=\textwidth]{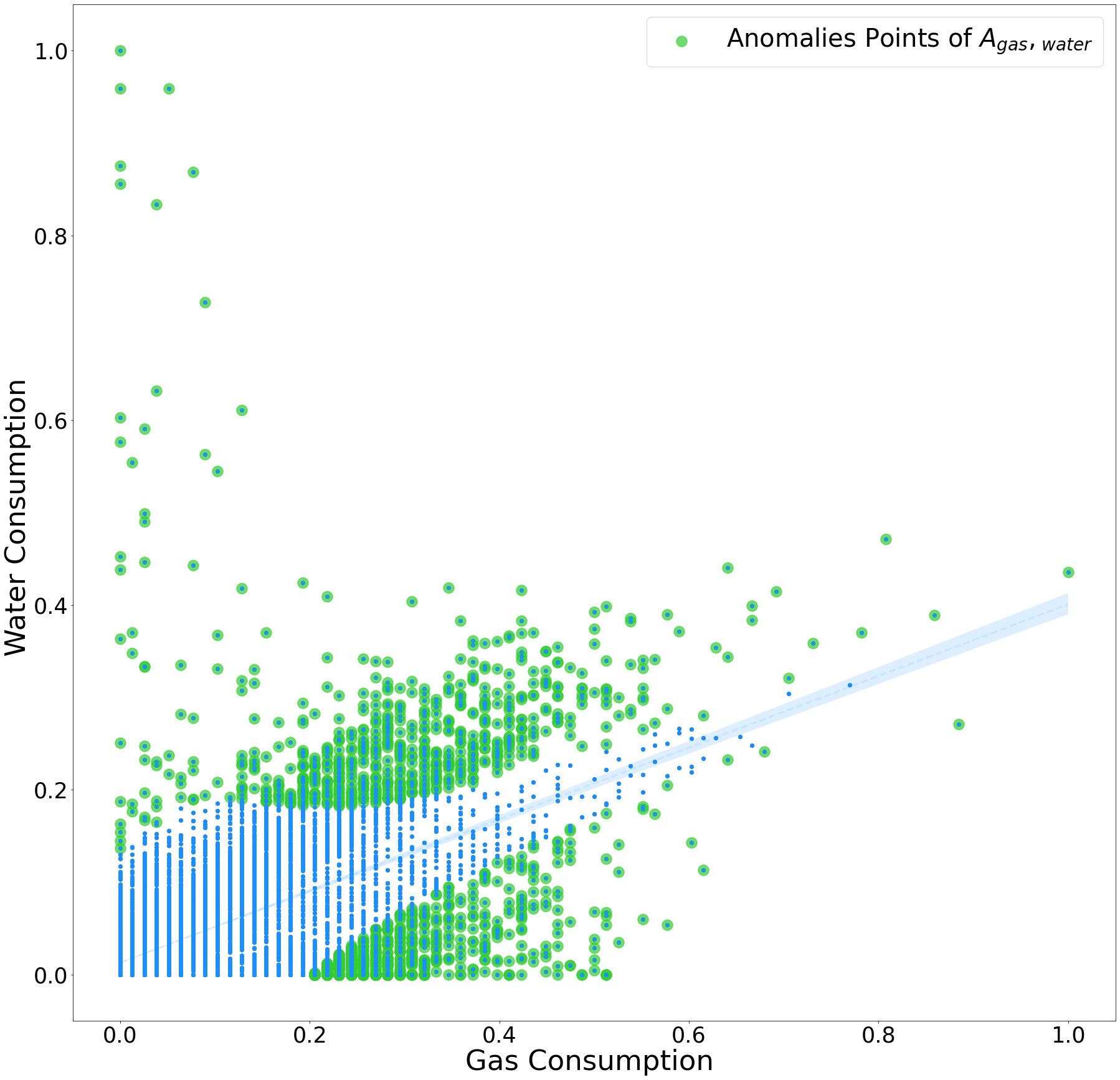}
         \caption{Linear regression of $CS_{water}$ on $CS_{gas}$}
         \label{fig5-b}
     \end{subfigure}
    \caption{Correlation-based Anomaly Definition.}
    \label{anomalydefinition}
\end{figure}

Fig.~\ref{anomalydefinition} shows an example of the correlation-based anomalies identified by the correlation between gas and water. In Fig.~\ref{fig5-a}, we define red-circled points as anomalies, $A_{water, gas}$, which are found by the linear regression model of $CS_{gas}$ on $CS_{water}$, indicating that some of them are regarded as normal data only if considering $CS_{gas}$. Similarly, in Fig.~\ref{fig5-b}, we define $A_{gas, water}$, found by the linear regression model of $CS_{water}$ on $CS_{gas}$.

Table~\ref{table1} presents the proportion of the final anomalies identified by our anomaly definition method. We adjust the portion of anomalies evenly across the datasets around 10$\sim$15\%. We note that anomalies are mostly defined by the correlation including anomalies defined by the individual sources, but a significant portion of anomalies can be detected only by the anomalies defined by the correlation. 

\begin{table*}[!t]
\centering
\renewcommand{\arraystretch}{1.5}
\caption{Proportion of defined anomalies on datasets.}
\begin{tabular}{lcccccc}

\Xhline{2\arrayrulewidth}
\multicolumn{1}{c}{\multirow{2}{*}{\textbf{Type}}} & \multicolumn{6}{c}{\textbf{Dataset}}                                     \\\cline{2-7}
\multicolumn{1}{c}{}                               & 3-ECK-2022 & 4-ECK-2022 & 5-ECK-2022 & 3-ECK-2021 & 4-ECK-2021 & AMPds2  \\\midrule
Individual-based (a)                               & 7.40\%     & 9.24\%     & 11.40\%    & 5.74\%     & 8.93\%     & 5.14\%  \\
Correlation-based (b)                              & 10.13\%    & 11.61\%    & 11.61\%    & 12.24\%    & 12.24\%    & 12.74\% \\
Intersection of (a) and (b)                        & 7.40\%     & 9.24\%     & 9.31\%     & 5.74\%     & 6.03\%     & 5.14\%  \\
(a) - (b)                                          & 0          & 0          & 2.09\%     & 0          & 2.09\%     & 0       \\
(b) - (a)                                          & 2.73\%     & 2.37\%     & 2.31\%     & 6.50\%     & 6.21\%     & 7.60\%  \\\hline
Final ratio                                        & 10.13\%    & 11.61\%    & 13.70\%    & 12.24\%    & 15.14\%    & 12.74\% \\\Xhline{2\arrayrulewidth}
\end{tabular}
\label{table1}
\end{table*}

\subsection{Correlation-Driven Multi-Level Multimodal Learning}
\label{sec:Correlation-Driven Multi-Level Multimodal Learning}
In this section, we propose a correlation-driven multi-level multimodal Learning model. In Section~\ref{sec:Two-Level Multimodal Learning Model}, we consider the case where three energy sources are correlated as the simplest case. Then, in Section~\ref{sec:Generalization of Two-Level Model}, we extend it to cover more energy sources that are correlated. Finally, in Section~\ref{sec:Three-Level Model to integrate Non-Correlated Energy Sources}, we present the final model that incorporates energy sources that are not correlated with the other energy sources. 

\subsubsection{Two-Level Multimodal Learning Model}
\label{sec:Two-Level Multimodal Learning Model}

In this section, we present a correlation-driven multi-level multimodal learning model for correlated energy sources. When it comes to fusing multiple energy sources in multimodal learning, we indicate that correlations between multiple energy sources are considerably variable as shown in Fig.~\ref{dataset-correlations}. This induces us to reflect them differentially according to the difference of their correlations, inspired by the fact that the correlation between multi-modalities represents the effects of fusing them~\cite{atrey2010multimodal}. To this end, we design a novel multimodal learning model by fusing them in multi-level based on the difference in the correlations between energy sources.

\begin{figure}[!t]
\centerline{\includegraphics[width=0.65\textwidth]{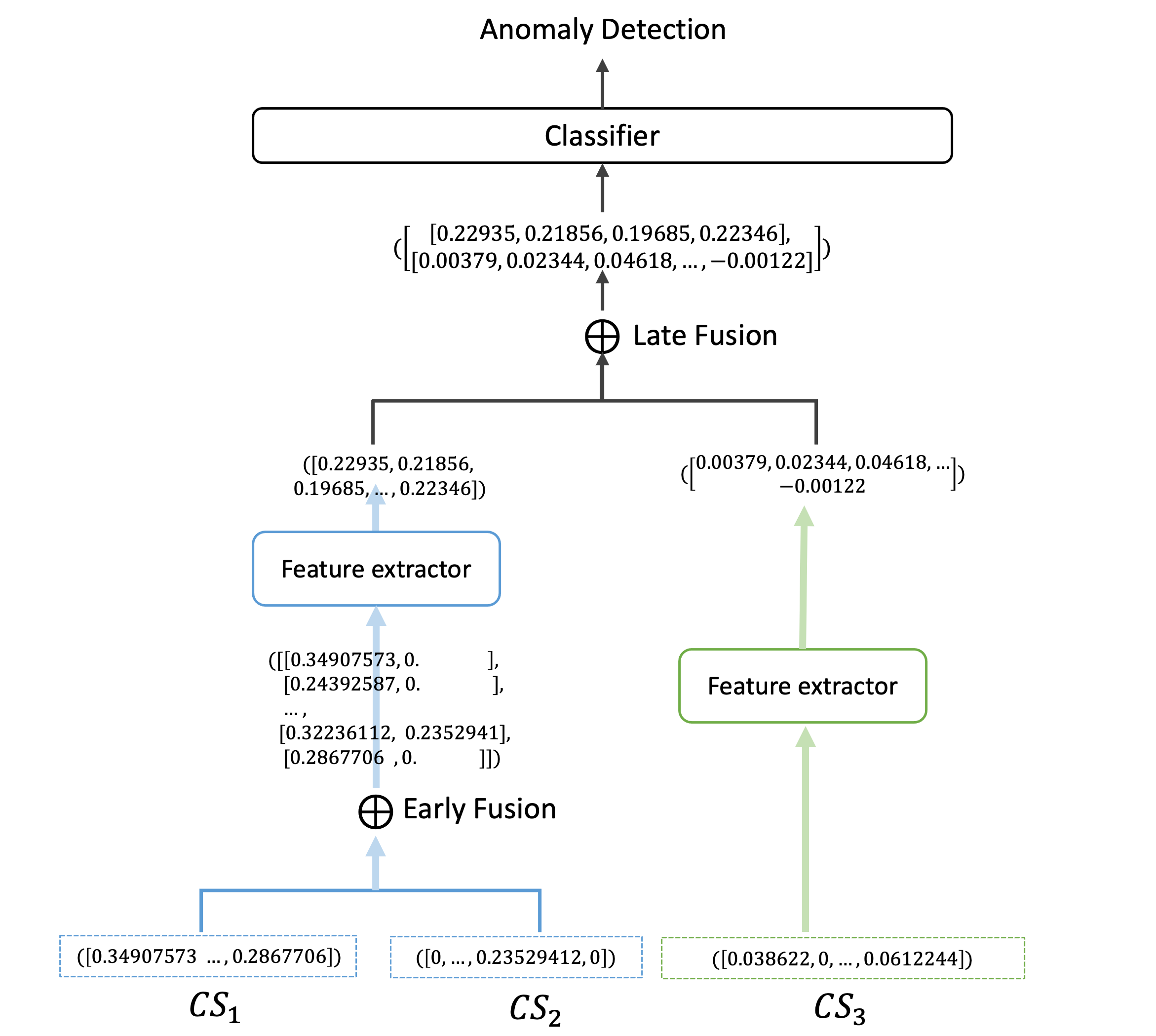}}
\caption{Correlation-driven two-level multimodal learning.}
\label{basic-model}
\end{figure}

Fig.~\ref{basic-model} describes the architecture of the proposed model. As the simplest case, we consider three kinds of energy sources to explain the main characteristics of the proposed model. A novel feature of the proposed model is that energy sources are placed at different levels according to the correlations between energy sources. Then, we apply the fusion methods~(i.e., early or late fusion) to the energy sources for each level. At the first level, the main strategy is that we integrate two energy sources that have the weakest correlation by early fusion to maximize integration effects. Early fusion is preferred to fuse features with distinct characteristics since it’s allowed to find potentially relevant relationships between features~\cite{baltruvsaitis2018multimodal, atrey2010multimodal}. Next, in the second level, we integrate the features of a strongly correlated energy source with the results of early fusion by late fusion. Hence, to preserve a strongly correlated energy source independently, we separately process it from the weakly correlated energy sources and extract feature representations individually.

The deep learning-based anomaly detection model logically consists of two components: 1)~feature extractor, which learns the representations of the raw input data, and 2)~classifier, which determines the final class of input data~(i.e., normal or abnormal)~based on the learned representations. For the feature extractor, we adopt unsupervised time-series representation learning that learns the representations in an unsupervised manner, and for the classifier, we apply the learned representations to the downstream tasks in a supervised manner~\cite{yang2022unsupervised, eldele2021time}. Hence, we extract enriched representations from multiple energy sources in the feature extractor, and then we exploit these learned representations for anomaly detection by learning the classifier with the outputs of the feature extractors.

\subsubsection{Generalization of Two-Level Model}
\label{sec:Generalization of Two-Level Model}

\begin{figure}[!t]
\centering
\centerline{\includegraphics[width=0.55\textwidth]{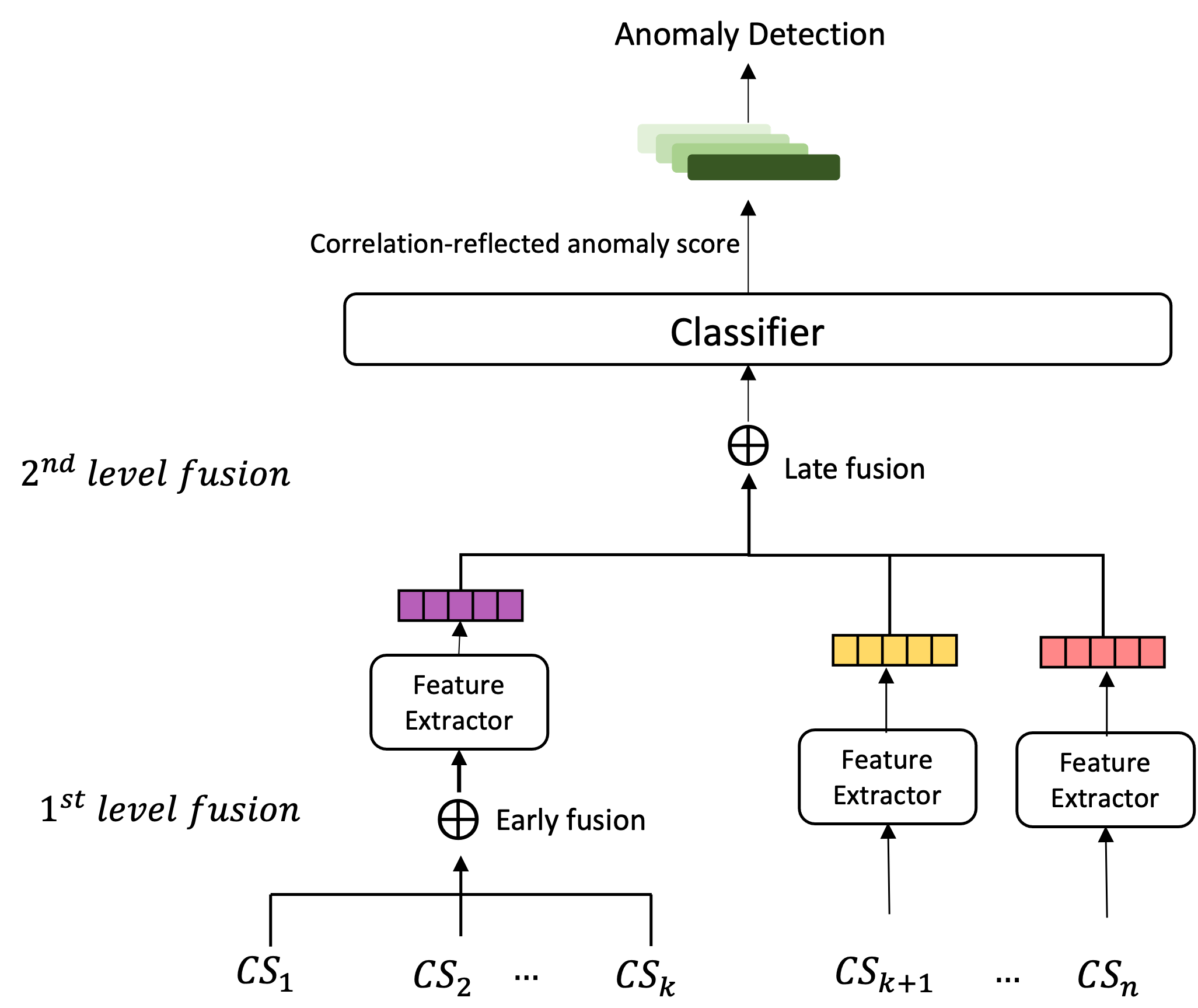}}
\caption{Generalization of Two-level Multimodal learning.}
\label{generalization-model}
\end{figure}

In this section, we generalize the proposed model in order to integrate more energy sources. Fig.~\ref{generalization-model} provides a generalized framework of our proposed model as the energy sources increase. We formalize our multimodal learning as follows. We define correlated energy sources as $CS=\{CS_1, ... , CS_n\}$, depending on whether each energy source is correlated with the other energy sources. Let $P=\{1,2, ..., p\}$ and $Q=\{p+1,\,p+2, ..., n\}$, where \textit{P} contains the energy sources for early fusion and \textit{Q} them for late fusion. We first select two correlated energy sources whose average correlation coefficients to the others are the weakest and add them into \textit{P} as the energy sources used for early fusion. Then, following Eq.~\ref{eq2}, we find an energy source with the minimum average of Pearson correlation coefficients with the other energy sources in \textit{P}. If its value by Eq.~\ref{eq2} is less than $Th_{correlated}$, we augment it to \textit{P}. This process will be repeated for the remaining energy sources. Finally, we add all the energy sources not in \textit{P} into \textit{Q}.

\begin{equation}\label{eq2}
\min_{k \in Q}\left\{\frac{\sum _{r \in P} Corr(CS_k, CS_r)}{n-(p+1)} \right\} 
\end{equation}

Considering the actual dataset, gas and power are in \textit{P} and water in \textit{Q} in AMPds2;~hot water and power in \textit{P}, and water in \textit{Q} on 3-ECK-2021;~gas and power in \textit{P}, and water in \textit{Q} on 3-ECK-2022;~hot water and power in \textit{P}, and water in \textit{Q} on 4-ECK-2021;~gas and hot water in \textit{P}, power, and water in \textit{Q} on 4-ECK-2022. In a similar way, we can determine the correlated and non-correlated energy sources and their orders for any dataset.

\subsubsection{Three-Level Model to integrate Non-Correlated Energy Sources}
\label{sec:Three-Level Model to integrate Non-Correlated Energy Sources}
\begin{figure}[!t]
\centering
\centerline{\includegraphics[width=0.55\textwidth]{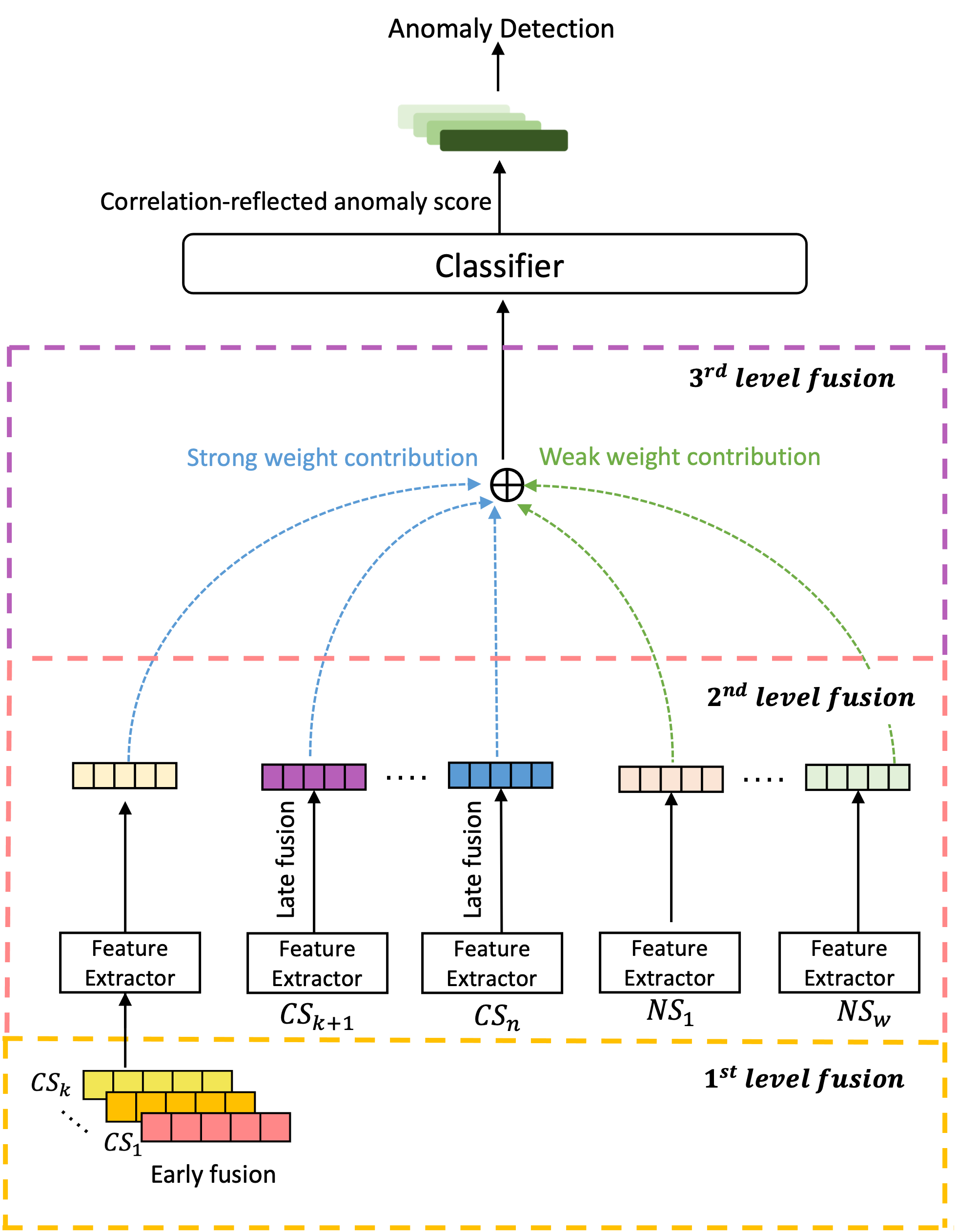}}
\caption{Three-level model integrating non-correlated energy sources.}
\label{non-model}
\end{figure}

We further extend the two-level multimodal learning model to integrate non-correlated energy sources. In the two-level model, all the energy sources are correlated with each other, but the strength is different. Although most energy sources have correlated with one another, some of them are not, e.g., heating in 5-ECK-2022. In this study, we devise a model to integrate non-correlated energy sources as shown in Fig.~\ref{non-model}. We denote non-correlated energy sources as $NS=\{NS_1, ... , NS_w\}$. For example, heating in 4-ECK-2021 and 5-ECK-2022 datasets is a non-correlated energy source. 

The key idea is to differentiate non-correlated energy sources from the correlated energy sources in the model. To achieve this goal, we learn the third-level layer in the model to have correlated and non-correlated energy source's differential weight contributions, which we denote \textit{correlation-based weight contribution}. We note that the distinguishing difference between the correlated and non-correlated energy sources is that the former generates the correlation-based anomalies, but the latter does not in the annotation. To utilize this characteristic, we train the third-level layer of the model by using the annotation in a supervised manner, differentiating the weight contributions between a group of correlated energy sources and a group of non-correlated energy sources. As a result, correlated energy sources will receive stronger weight contributions than non-correlated energy sources by learning the correlation-based anomalies.

According to the correlation-based weight contribution, the importance of the correlated and non-correlated energy source in the third-level layer of our model is adjusted as the result of supervised learning while they are concatenated together. We verify its impact by measuring the differential weight contributions between the source of the correlated and non-correlated. For this, we measure the SHAP value (SHapley Addictive exPlanations) of each energy source~\cite{lundberg2017unified} to represent its weight contribution to the results. Here, we use the 5-ECK-2022 dataset including two weakly correlated sources, gas and hot water, two strongly correlated sources, power and water, and one non-correlated energy source, heating.

\begin{figure}[!t]
\captionsetup[subfigure]{justification=centering}
     \centering
     \begin{subfigure}[h]{0.48\textwidth}
         \includegraphics[width=\textwidth]{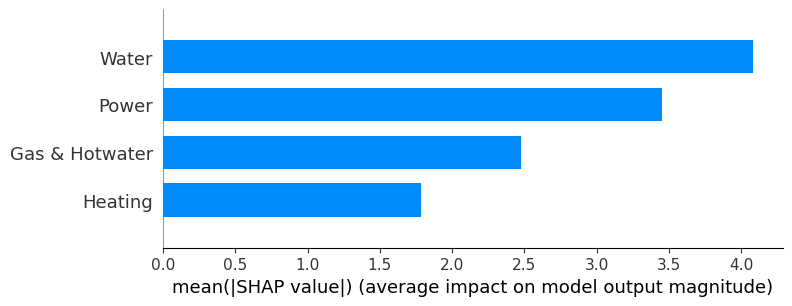}
         \caption{The proposed model.}
         \label{shap-a}
     \end{subfigure} 
     \begin{subfigure}[h]{0.48\textwidth}
         \includegraphics[width=\textwidth]{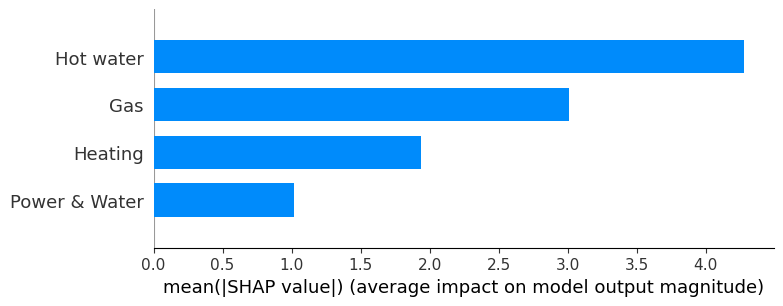}
         \caption{Replacing strongly correlated energy sources with weakly correlated energy sources.}
         \label{shap-b}
     \end{subfigure}
     \hfill
     \begin{subfigure}[h]{0.48\textwidth}
         \includegraphics[width=\textwidth]{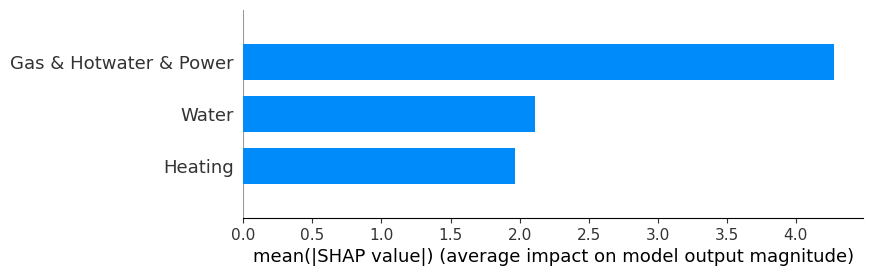}
         \caption{Adding a strongly correlated energy source to early fusion.}
         \label{shap-c}
     \end{subfigure}
    \hfill 
     \begin{subfigure}[h]{0.48\textwidth}
         \includegraphics[width=\textwidth]{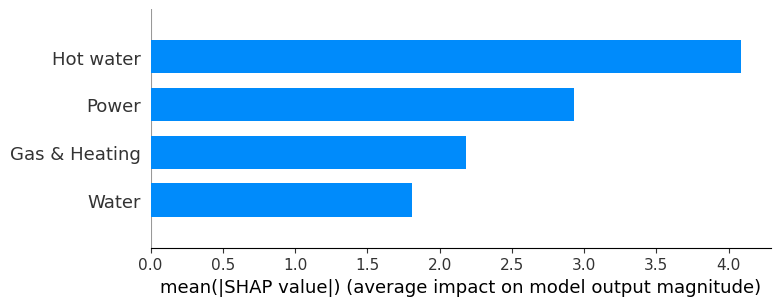}
         \caption{Exploiting a non-correlated energy source as the weakest correlated energy source.}
         \label{shap-d}
     \end{subfigure}
        \caption{Weight contribution analysis between the proposed model and comparisons on 5-ECK-2022.}
        \label{shap}
\end{figure}

Fig.~\ref{shap} compares the SHAP values of the energy sources in the third level of our proposed model~(Fig.~\ref{shap-a})~with the following three comparisons: 1)~Replacing the strongly correlated energy sources with the weakly correlated energy sources~(Fig.~\ref{shap-b}), 2)~adding a strongly correlated energy source to early fusion (Fig.~\ref{shap-c}), and 3)~regarding a non-correlated energy source as the weakest correlated energy source~(Fig.~\ref{shap-d}). We note that energy sources contribute to the results in the order of strongly correlated, weakly correlated, and non-correlated energy sources in the proposed model as we intended, meanwhile the comparisons are not. In fact, these salient points are relevant to significant performance improvements of the proposed model compared to the comparisons, which will be discussed in Section~\ref{sec: Impact of Correlation-based Weight Contribution}.

%% file: ch6_performance_evaluation.tex
\section{Performance Evaluation} \label{experiments}

\subsection{Experimental Environments and Methods}
Through extensive experiments, we compare our proposed method with the existing models including 1) the representative multimodal learning models, i.e., early and late fusion~\cite{baltruvsaitis2018multimodal, atrey2010multimodal}, 2) recent models in time-series anomaly detection, i.e., USAD~\cite{audibert2020usad} and LSTM-AE~\cite{malhotra2016lstm}, and TAnoGAN~\cite{bashar2020tanogan}. As explained in Section 5.2.1, we learn a feature extractor with an unsupervised manner to effectively learn the representations from normal data and a classifier with a supervised manner using the outputs of the learned feature extractor along with the annotations. Regarding training and evaluation, we divide the entire dataset into training and test dataset with a portion of 7:3. Then, we use only normal data in the training dataset to learn a feature extractor, but for a classifier, we use both abnormal and normal data in the training dataset. We note that we can employ any existing anomaly detection models as the feature extractor and classifier of our multimodal learning. Therefore, we focus on the comparison of relative performance between our modal and the baseline model used for the feature extractor. In the experiments, we adopted USAD~\cite{audibert2020usad}, one of the state-of-the-art models for unsupervised anomaly detection on multivariate time series, as a feature extractor, and replaced it with LSTM-AE~\cite{malhotra2016lstm} so as to show that our model consistently outperforms baseline models used for the feature extractor. For fair comparisons, we consistently used LightGBM~\cite{ke2017lightgbm} as a classifier for all models.

Table~\ref{table2} shows hyperparameters that are consistently used for both our model and the comparison models. To guarantee performance comparison with anomaly detection models, we follow the existing approach~\cite{audibert2020usad} that finds the best anomaly threshold showing the highest F1 score for each model by evaluating its performance within a feasible range of anomaly thresholds. 
\begin{table}
\setlength{\tabcolsep}{8pt}
\centering
\caption{Common hyperparameters for the models.}
\label{enzymatic cocktails}
\begin{tabular}{@{}ccc@{}}
\toprule
\textbf{Parameters}  & {}         & \textbf{Values}                                              \\ \hline\hline
\multirow{5}{*}{Feature extractor}& \multicolumn{1}{l|}{Batch size} & 1020 \\
                   & \multicolumn{1}{l|}{Epochs}            & 100  \\
                    & \multicolumn{1}{l|}{Sequence length}            & 12  \\
                    & \multicolumn{1}{l|}{Learning rate}            & 0.01  \\
                    & \multicolumn{1}{l|}{Optimizer}            & Adam  \\\cmidrule(r){1-2}\cmidrule(r){2-3}

\multirow{3}{*}{Classifier}& \multicolumn{1}{c|}{Number of estimators}                & 400 \\
                   & \multicolumn{1}{l|}{loss function}            & logloss  \\
                   & \multicolumn{1}{l|}{boosting type} &  decision tree \\ \bottomrule
\end{tabular}
\label{table2}
\end{table}

In the experiments, we use four evaluation metrics: (1) AUROC, (2) Macro F1-score (M-F1), (3) Weighted F1-score (W-F1), and (4) F1 score. Out of them, we use the F1 score as a representative comparison metric to compare the extent of performance improvement between experiments since it is the most commonly used metric. We performed all experiments on a machine equipped with NVIDIA RTX A4000 and Intel(R) Xeon(R) Silver 4210R CPU @ 2.40GHz with 32GB memory, running on Ubuntu 18.04. We used Python(version 3.8.11) and PyTorch(version 1.11.0) to implement the proposed model and the comparison models.

\subsection{Experimental Results}
\subsubsection{Effects of Correlation-Driven Feature Placement}

We verify the effectiveness of our novel characteristic that places features based on their correlations in multimodal learning. Here, we used AMPds2 because the correlations of two energy sources are clearly weaker than the other.  Table~\ref{table3} shows performance comparison when we use the same architecture of the multi-level model, but place the features at different levels. We note that the proposed model consistently outperforms the comparison models. Hence, in AMPds2, when the energy sources having weaker correlations with the other are integrated by early fusion, it outperforms other cases. 
\begin{table*}[!t]
\caption{Performance comparison to validate correlation-driven feature placement in multimodal learning.}
\centering
\setlength{\tabcolsep}{4pt}
\renewcommand{\arraystretch}{1.5}
\begin{tabular}{ccccccc}
\toprule
\multirow{1}{*}                                               & 
\multicolumn{2}{c}{Fusion}                 & \multicolumn{4}{c}{AMPds2}                                            \\\cmidrule(rl){2-3}\cmidrule(rl){4-7}
                                                                       & 1st level                 & 2nd level      & AUROC           & M-F1            & W-F1            & F1              \\\hline \hline
\multirow{1}{*}{\begin{tabular}[c]{@{}c@{}}\end{tabular}}                                                & Water \&Gas                & Power          & 0.8328          & 0.8340          & 0.8953          & 0.9350          \\
                                                                       & Power \&Water              & Gas            & 0.8419          & 0.8376          & 0.8967          & 0.9350          \\
                                                                       & Gas \&Power     & Water & \textbf{0.9064} & \textbf{0.8772} & \textbf{0.9194} & \textbf{0.9468} \\\bottomrule
\end{tabular}
\label{table3}
\end{table*}

\subsubsection{Performance Comparison with Correlated Three Energy Sources }
\label{Sec: Performance Comparison with Three Energy Sources}

\begin{table*}[!t]
\setlength{\tabcolsep}{3.5pt}
\centering
\renewcommand{\arraystretch}{1.4}
\caption{Performance comparison on the datasets with three energy sources.}

\begin{tabular}{lcccccccccccc}
\Xhline{2\arrayrulewidth}

\multicolumn{1}{c}{\multirow{2}{*}{Methods}} & \multicolumn{4}{c}{AMPds2}                                            & \multicolumn{4}{c}{3-ECK-2022}                                        & \multicolumn{4}{c}{3-ECK-2021}                                        \\\cmidrule(rl){2-5}\cmidrule(rl){6-9}\cmidrule(rl){10-13} 
\multicolumn{1}{c}{}                         & AUROC           & M-F1            & W-F1            & F1              & AUROC           & M-F1            & W-F1            & F1              & AUROC           & M-F1            & W-F1            & F1              \\\hline \hline
USAD                                         & 0.9059          & \textbf{0.9021} & 0.9063          & 0.9213          & 0.8192          & 0.8147          & 0.8192          & 0.8407          & \textbf{0.8882} & \textbf{0.8758} & 0.8896          & 0.9148          \\
LSTM-AE                                      & 0.7355          & 0.7476          & 0.7709          & 0.8544          & 0.5729          & 0.5686          & 0.5761          & 0.6114          & 0.7616          & 0.7918          & 0.8287          & 0.8967          \\
TAnoGAN                                      & 0.7262          & 0.7804          & 0.8776          & 0.9384          & 0.7146          & 0.6647          & 0.7336          & 0.7929          & 0.8856          & 0.8275          & 0.8711          & 0.9081          \\
AVG.                                         & 0.7892          & 0.8100          & 0.8516          & 0.9047          & 0.7022          & 0.6827          & 0.7096          & 0.7483          & 0.8451          & 0.8317          & 0.8631          & 0.9065          \\\hline\hline
Early fusion                                 & 0.8405          & 0.8435          & 0.9015          & 0.9392          & 0.8281          & \textbf{0.8234} & 0.8337          & 0.8639          & 0.6756          & 0.7101          & 0.8657          & 0.9319          \\
Late fusion                                  & 0.8690          & 0.8569          & 0.9080         & 0.9412          & \textbf{0.8287} & 0.8189          & 0.8282          & 0.8555          & 0.7758          & 0.7667          & 0.8795          & 0.9275          \\
AVG.                                         & 0.8548          & 0.8502          & 0.9048          & 0.8402          & 0.8284          & 0.8212          & 0.8310          & 0.8597          & 0.7257          & 0.7384          & 0.8726          & 0.9297          \\\hline\hline
\textbf{Ours}                                & \textbf{0.9064} & 0.8772          & \textbf{0.9194} & \textbf{0.9468} & 0.8204          & 0.8175          & \textbf{0.8395} & \textbf{0.8799} & 0.7788          & 0.8237          & \textbf{0.9168} & \textbf{0.9564} \\\bottomrule
\end{tabular}
\label{3data-performance}
\end{table*}

To verify the effectiveness of the proposed correlation-based model design, we first compare the performance between our model and the comparison models on datasets with three energy sources where all of which are correlated with each other. Table~\ref{3data-performance} presents the results of the performance comparison between the proposed model and five comparisons. In this regard, due to the fact that the performance varies by evaluation metrics, we use the F1 score as the representative metric to compare the performance between models. Overall, our model generally outperforms the comparison models for all datasets. Specifically, our model improves the performance of the existing multimodal learning models by 0.56$\sim$2.89\%\ and that of time series anomaly detection methods by 0.84$\sim$26.85\%\ based on the F1 score. We note that the performance trends become clear and more consistent as more energy sources are integrated into the upcoming experimental results regardless of the evaluation metrics, showing the structural advantages of our model.

\subsubsection{Performance Comparison with Correlated Four Energy Sources }
To verify the scalability of the proposed model for multiple energy sources, we use more than three energy sources and compare the results with the previous results on datasets with three energy sources. In this section, we focused on correlated energy sources, 4-ECK-2022. Table~\ref{4data-performance} presents the experimental results to compare the performance between our model and the comparison models. We indicate that our model significantly outperforms the comparison models. Specifically, it improves the existing multimodal learning models by 12.48$\sim$14.99\%\ and time series anomaly detection methods by 6.41$\sim$25.32\%\ based on the F1 score. More importantly, we note that as the energy sources increase, further performance improvements of our model are achieved, showing the effectiveness and scalability of the proposed model. Specifically, the performance of our method improves the existing models by 13.73$\sim$14.88\%\ based on the average F1 score on 4-ECK-2022, whereas it was 2.02$\sim$13.16\%\ on 3-ECK-2022 in Section~\ref{Sec: Performance Comparison with Three Energy Sources}, which is a subset of 4-ECK-2022.


\begin{table*}[!t]
\renewcommand{\arraystretch}{1.5}
\centering
\caption{Performance comparison on the dataset with correlated four energy sources.}

\begin{tabular}{lllll}
\Xhline{2\arrayrulewidth}
\multicolumn{1}{c}{\multirow{2}{*}{Methods}} & \multicolumn{4}{c}{4-ECK-2022}                                                                                    \\ 
\cmidrule(rl){2-5}
\multicolumn{1}{c}{}                         & \multicolumn{1}{c}{AUROC}  & \multicolumn{1}{c}{M-F1}   & \multicolumn{1}{c}{W-F1}   & \multicolumn{1}{c}{F1}     \\ \hline\hline
USAD~\cite{audibert2020usad}                                         &0.8597 & \multicolumn{1}{c}{0.8199} & \multicolumn{1}{c}{0.8648} & \multicolumn{1}{c}{0.9031} \\
LSTM-AE~\cite{malhotra2016lstm}                                      &0.6443 & \multicolumn{1}{c}{0.6217} & \multicolumn{1}{c}{0.6567} & \multicolumn{1}{c}{0.7140} \\
TAnoGAN~\cite{bashar2020tanogan}                                      & 0.7850                     & 0.7588                     & 0.7921                     & 0.8382                     \\
AVG.                                         & 0.7630                     & 0.7335                     & 0.7712                     & 0.8184                     \\\hline\hline
Early fusion                                 & 0.7129                     & 0.7205                     & 0.7689                     & 0.8424                     \\
Late fusion                                  & 0.4794                     & 0.4272                     & 0.6094                     & 0.8173                     \\
AVG.                                         & 0.5962                     & 0.5739                     & 0.6892                     & 0.8299                     \\\hline\hline
\textbf{Ours}                                         & \textbf{0.9505}                     & \textbf{0.9464}                     & \textbf{0.9547}                     & \textbf{0.9672}                    \\\bottomrule
\end{tabular}
\label{4data-performance}
\end{table*}

\subsubsection{Performance Comparison with Non-Correlated Energy Sources }
In this section, we compare the performance between the proposed model and the comparison models using the datasets, i.e., 4-ECK-2021 and 5-ECK-2022, including a non-correlated energy source. Heating is a non-correlated source in both datasets, and the others are correlated energy sources. Table~\ref{non-correlated-performance} shows the performance comparison for the datasets. This result indicates that our model outperforms the existing multimodal learning models by 0.41$\sim$6.69\%\ and time series anomaly detection methods by 5.81$\sim$29.24\%\ based on the F1 score.

By comparing the performance across the datasets, we observe two key strengths of the proposed model. First, our model effectively integrates non-correlated energy sources with further performance improvement compared to the case where all the energy sources are correlated. Specifically, our model improves the performance of the comparisons by

3.25$\sim$23.36\%\ based on the average F1 score in the case of 4-ECK-2021 and 5-ECK-2022 where non-correlated energy sources are added in the datasets, whereas it improves by

2.67$\sim$14.88\%\ in the case of 3-ECK-2021 and 4-ECK-2022, which only excludes the non-correlated energy source from 4-ECK-2021 and 5-ECK-2022, respectively. Second, we confirm the scalability of the proposed model as energy sources increase. Specifically, our method improves the performance of the comparison models by 4.77$\sim$23.36\%\ based on the average F1-score in 5-ECK-2022 consisting of five energy sources while it improves by 3.25$\sim$8.22\%\ in 4-ECK-2021 consisting of four energy sources.

\begin{table*}[!t]
\setlength{\tabcolsep}{10pt}
\centering
\caption{Performance comparison on datasets including non-correlated sources.}
\renewcommand{\arraystretch}{1.5}
\begin{tabular}{lcccccccc}
\toprule
\multicolumn{1}{c}{\multirow{2}{*}{Methods}} & \multicolumn{4}{c}{4-ECK-2021}                               & \multicolumn{4}{c}{5-ECK-2022}                               \\\cmidrule(rl){2-5}\cmidrule(rl){6-9}
\multicolumn{1}{c}{}                                  & AUROC           & M-F1            & W-F1            & F1              & AUROC           & M-F1            & W-F1            & F1              \\\hline\hline
USAD                                                  & \textbf{0.8845} & \textbf{0.8680} & 0.8866          & 0.9146          & 0.7600          & 0.7112          & 0.7751          & 0.8295          \\
LSTM-AE                                               & 0.6384          & 0.6491          & 0.7337          & 0.8636          & 0.5319          & 0.5213          & 0.6199          & 0.7143          \\
TAnoGAN                                               & 0.7266          & 0.7584          & 0.8116          & 0.8932          & 0.6452          & 0.6101          & 0.6380          & 0.6837          \\
AVG.                                                  & 0.7498          & 0.7585          & 0.8106          & 0.8905          & 0.6457          & 0.6142          & 0.6777          & 0.7425          \\\hline\hline
Early fusion                                          & 0.7774          & 0.8127          & 0.8518          & 0.9118          & 0.5676          & 0.5799          & 0.8008          & 0.9092          \\
Late fusion                                           & 0.8661          & 0.8456          & 0.9451          & 0.9686          & 0.8388          & 0.8515          & 0.9120          & 0.9476          \\
AVG.                                                  & 0.8218          & 0.8292          & 0.8985          & 0.9402          & 0.7032          & 0.7157          & 0.8564          & 0.9284          \\\hline\hline
\textbf{Ours}                                                  & 0.8016          & 0.8376          & \textbf{0.9470} & \textbf{0.9727} & \textbf{0.9269} & \textbf{0.9274} & \textbf{0.9601} & \textbf{0.9761}\\\bottomrule
\end{tabular}
\label{non-correlated-performance}
\end{table*}

\subsubsection{Effects of Correlation-based Weight Contribution }
\label{sec: Impact of Correlation-based Weight Contribution}
We investigate the effectiveness of correlation-based weight contribution on the 5-ECK-2022 dataset. We design three the comparison models in the same way described in Section~\ref{sec:Three-Level Model to integrate Non-Correlated Energy Sources}. Table~\ref{ablation} describes the results of the performance evaluation. Our proposed model explicitly outperforms all the comparisons, showing the strength of our model by effectively dealing with multiple energy sources including even non-correlated sources. 

\begin{table*}[!t]
\setlength{\tabcolsep}{8pt}
\caption{Performance comparison to show the effects of correlation-based weight contribution.}
\renewcommand{\arraystretch}{1.5}
\centering
\begin{tabular}{ccccc}
\toprule
\multirow{2}{*}{Model} & \multicolumn{4}{c}{5-ECK-2022}                                            \\\cmidrule(rl){2-5}
                       & AUROC           & M-F1            & W-F1            & F1              \\\hline\hline
Case1                  & 0.8074          & 0.8359          & 0.8659          & 0.9149          \\
Case2                  & 0.8717          & 0.7066          & 0.8336          & 0.8959          \\
Case3                  & 0.8644          & 0.8605          & 0.8957          & 0.9300          \\\hline\hline
\textbf{Ours}          & \textbf{0.9269} & \textbf{0.9274} & \textbf{0.9601} & \textbf{0.9761}\\\bottomrule
\end{tabular}
\label{ablation}
\end{table*}

\subsubsection{Performance Comparison on Seasonal Datasets }
\begin{figure*}[!t]
\captionsetup[subfigure]{justification=centering}
     \centering
     \begin{subfigure}[h]{0.4\textwidth}
         \centering
         \includegraphics[width=\textwidth]{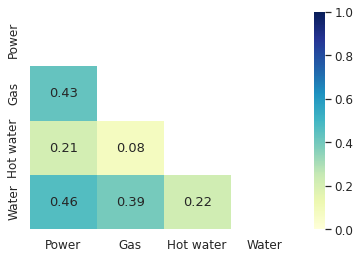}
         \caption{Winter}
         \label{fig10-a}
     \end{subfigure} 
     \begin{subfigure}[h]{0.4\textwidth}
        \centering
         \includegraphics[width=\textwidth]{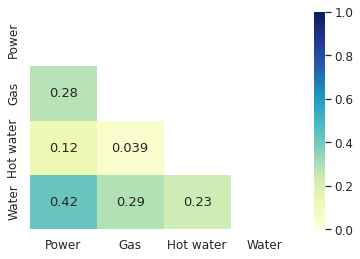}
         \caption{Summer}
         \label{fig10-b}
     \end{subfigure}
    \caption{Pearson correlation coefficients of seasonal data.}
    \label{seasonal}
\end{figure*}

  In this section, we measure the performance of the proposed model considering seasonal variations. We note that the correlation between energy sources varies with the seasons. As shown in Fig.~\ref{seasonal}, hot water and gas are higher correlated with power in winter than in summer. This stems from the fact that, in winter, people tend to consume hot water and gas while staying at home.

Table~\ref{seasonal-performance} demonstrates the performance of the seasonal datasets of 4-ECK-2022. As the correlation between energy sources varies by seasons, our model reflects the order of energy sources to be integrated into the model based on the correlation. Specifically, in winter, our model first integrates gas and hot water by early fusion, then water and power subsequently incorporated by late fusion. In contrast, in summer, it integrates gas and hot water by early fusion and then integrates power and water by late fusion. The results show that our proposed model consistently outperforms all the comparison models across winter and summer datasets, demonstrating the superiority of our model.

\begin{table*}[h]
    \begin{subtable}[h]{0.45\textwidth}
        \centering
        \caption{Winter}
        \renewcommand{\arraystretch}{1.3}
        \begin{tabular}{lcccc}
        \toprule
        \multicolumn{1}{c}{\multirow{2}{*}{Methods}} & \multicolumn{4}{c}{4-ECK-2022}                                        \\\cmidrule(rl){2-5}
        \multicolumn{1}{c}{}                         & AUROC           & M-F1            & W-F1            & F1              \\\hline\hline
        USAD                                         & 0.7561          & 0.7276          & 0.7647          & 0.8149          \\
        LSTM-AE                                      & 0.6412          & 0.6130          & 0.6569          & 0.7174          \\
        TAnoGAN                                      & 0.7102          & 0.6775          & 0.7170          & 0.7705          \\
        Early fusion                                 & 0.7079          & 0.7130          & 0.8008          & 0.8740          \\
        Late fusion                                  & 0.8993          & 0.8831          & 0.9023          & 0.9282          \\\hline\hline
        \textbf{Ours}                                & \textbf{0.9388} & \textbf{0.9366} & \textbf{0.9553} & \textbf{0.9709}\\\bottomrule
        \end{tabular}

       \label{winter}
    \end{subtable}
    \hfill
    \begin{subtable}[h]{0.45\textwidth}
        \centering
        \renewcommand{\arraystretch}{1.3}
        \caption{Summer}
        \begin{tabular}{lcccc}
        \toprule
        \multicolumn{1}{c}{\multirow{2}{*}{Methods}} & \multicolumn{4}{c}{4-ECK-2022}                                            \\\cmidrule(rl){2-5}
        \multicolumn{1}{c}{}                         & AUROC           & M-F1            & W-F1            & F1              \\\hline\hline
        USAD                                         & 0.8656          & 0.8275          & 0.8711          & 0.9081          \\
        LSTM-AE                                      & 0.7085          & 0.6587          & 0.7283          & 0.7882          \\
        TAnoGAN                                      & 0.6962          & 0.7146          & 0.8088          & 0.8873          \\
        Early fusion                                 & 0.7975          & 0.7859          & 0.8212          & 0.8689          \\
        Late fusion                                  & 0.8984          & 0.8819          & 0.9012          & 0.9273          \\\hline\hline
        \textbf{Ours}    & \textbf{0.9187} & \textbf{0.8889} & \textbf{0.9056} & \textbf{0.9281} \\\bottomrule
        
        \end{tabular}

        \label{summer}
     \end{subtable}
     \caption{Performance comparison on seasonal datasets.}
     \label{seasonal-performance}
\end{table*}

To verify the effectiveness of our fusion strategy in the model, we perform experiments by varying fusion strategies in the same architecture on 4-ECK-2022. First, for the dataset labeled based on correlation in all-season data, we construct two models in a different fusion strategy, the correlation of which is subject to 1) all seasons and 2) winter data. Here, the former is the proposed model, and the latter is the comparison model. Second, for the dataset labeled based on correlation in winter data, we construct two models in a different fusion strategy, the correlation of which is subject to 1) all seasons and 2) winter data. Here, the latter is the proposed model, and the former is the comparison model. Table~\ref{fusionstrategies} shows the experimental results, clearly showing that the proposed model outperforms the comparison models in both cases. This confirms the effectiveness of the proposed fusion strategy in the multimodal learning model.

\begin{table*}[!t]
\setlength{\tabcolsep}{3pt}
\centering
\renewcommand{\arraystretch}{1}
\caption{Performance comparison between different fusion strategies on seasonal datasets.}
\begin{tabular}{ccccccl}
\toprule
\multirow{2}{*}{Label Criteria}                                                     & \multirow{1}{*}{\begin{tabular}[c]{@{}c@{}}Correlation-based\\ Fusion strategy\end{tabular}} & \multicolumn{4}{c}{4-ECK-2022}                                        \\\cmidrule(rl){3-6}
                                                                                    &                                                                                              & AUROC           & M-F1            & W-F1            & F1              \\\hline\hline
\multirow{3}{*}{\begin{tabular}[c]{@{}c@{}}All seasons\\ (4-ECK-2021)\end{tabular}} & Winter                                                                                       & 0.9346          & 0.9329          & 0.9528          & 0.9693          \\
                                                                                    & \begin{tabular}[c]{@{}c@{}}All seasons \\ (Proposed)\end{tabular}                   & \textbf{0.9388} & \textbf{0.9366} & \textbf{0.9553} & \textbf{0.9709} \\\midrule
\multirow{3}{*}{\begin{tabular}[c]{@{}c@{}}Winter\\ (4-ECK-2021)\end{tabular}}      & All seasons                                                                                  & 0.9135          & 0.9161          & 0.9346          & 0.9557          \\
                                                                                    & \begin{tabular}[c]{@{}c@{}}Winter\\ (Proposed)\end{tabular}                         & \textbf{0.9505} & \textbf{0.9545} & \textbf{0.9464} & \textbf{0.9547}\\\bottomrule
\end{tabular}
\label{fusionstrategies}
\end{table*}

\subsubsection{Variations of Feature Extractors}
In this section, we replace USAD with LSTM-AE~\cite{malhotra2016lstm} for the feature extractor to show the proposed model’s structural advantages regardless of the model used for the feature extractor. Here, we used 4-ECK-2022 for this experiment. Table~\ref{ablation2} shows that our model based on LSTM-AE significantly outperforms LSTM-AE, similarly as in USAD, confirmed the structural advantages of our model.

\begin{table*}[!t]
\centering
\setlength{\tabcolsep}{4pt}
\renewcommand{\arraystretch}{1.5}
\caption{Performance of the proposed model on different feature extractors.}
\begin{tabular}{lcccc}
\toprule
\multicolumn{1}{c}{\multirow{2}{*}{Models}} & \multicolumn{4}{c}{4-ECK-2022}        \\\cmidrule(rl){2-5}
\multicolumn{1}{c}{}                        & AUROC  & M-F1   & W-F1   & F1     \\\hline\hline
USAD                                        & 0.8597 & 0.8199 & 0.8648 & 0.9031 \\
Proposed Model (USAD)                       & \textbf{0.9505} & \textbf{0.9464} & \textbf{0.9547} & \textbf{0.9672} \\ \midrule
LSTM-AE                                     & 0.6443 & 0.6217 & 0.6567 & 0.7140 \\
Proposed Model (LSTM-AE)                    & \textbf{0.7547} & \textbf{0.7799} & \textbf{0.8182} & \textbf{0.8867}\\\bottomrule
\end{tabular}
\label{ablation2}
\end{table*}

%% file: ch7_conclusion.tex
\section{Conclusions} \label{conclusion}

In this paper, we proposed a new multimodal learning model to effectively integrate multiple energy sources. Specifically, based on the strength of the correlations among energy sources, we presented an effective strategy to place energy sources in multi-level along with different fusing methods~(i.e., early fusion and late fusion). The main strategy was that we integrated weakly correlated energy sources using early fusion in the first-level layer of the model. Then, we integrated strongly correlated energy sources with the previous results of early fusion at the second-level layer. We also generalized the proposed scheme to cover more energy sources in the two-level multimodal learning model. For this, we extended the proposed model to integrate not only correlated energy sources but also non-correlated energy sources with a three-level multimodal learning model. In the three-level layer, we differentiated the weight contribution between correlated and non-correlated energy sources by learning the layer with the annotated datasets defined by considering the correlation in a supervised manner. Through the experiments using three real-world datasets, we clearly showed the effectiveness of the placement of energy sources in a multi-level model depending on their correlation. We indicated that our proposed model significantly outperformed the comparison models. In particular, we also proved that our proposed model further improved the performance as the source of the correlated and non-correlated energy increase.

In this paper, since we were not able to utilize ground truth anomaly datasets due to the lack of annotations in energy consumption datasets, we presented a new method that defines anomalous energy consumption including anomalies based on the correlation between energy sources. As a result, we can define an abnormal consumption considering the correlation between energy sources that cannot be found by an individual energy consumption alone. The salient point of the proposed model is that the fusion strategy of the model can be generalized to any type of data including energy sources because our model incorporates not only correlated but also non-correlated sources. Therefore, as a further study, we plan to extend the proposed idea to other domain problems dealing with multiple features by investigating the differential correlation between them.